\pdfoutput=1

\documentclass[11pt]{article}

\usepackage{EACL2023}

\usepackage{times}
\usepackage{latexsym}
\usepackage{caption}
\usepackage{subcaption}
\usepackage{adjustbox}
\usepackage{amsmath}
\usepackage{longtable}
\usepackage{xcolor}

\usepackage[T1]{fontenc}

\usepackage[utf8]{inputenc}

\usepackage{microtype}
\usepackage{inconsolata}

\title{How do decoding algorithms distribute information in dialogue responses?}

\author{Saranya Venkatraman \\ Pennsylvania State University \\ \texttt{saranyav@psu.edu}\\ \And He He \\ New York University \\ \texttt{hhe@nyu.edu}\\ \And David Reitter \\ Google Research \\ \texttt{reitter@google.com} \\}

\begin{document}
\maketitle
\begin{abstract}
    Humans tend to follow the Uniform Information Density (UID) principle by distributing information evenly in utterances. We study if decoding algorithms implicitly follow this UID principle, and under what conditions adherence to UID might be desirable for dialogue generation. We generate responses using different decoding algorithms with GPT-2 on the Persona-Chat dataset and collect human judgments on their quality using Amazon Mechanical Turk. We find that (i) surprisingly, model-generated responses follow the UID principle to a greater extent than human responses, and (ii) decoding algorithms that promote UID do not generate higher-quality responses. Instead, when we control for surprisal, non-uniformity of information density correlates with the quality of responses with very low/high surprisal. Our findings indicate that encouraging non-uniform responses is a potential solution to the ``likelihood trap'' problem (quality degradation in very high-likelihood text). Our dataset containing multiple candidate responses per dialog history along with human-annotated quality ratings is available at: \url{https://huggingface.co/datasets/saranya132/dialog_uid_gpt2}.

\end{abstract}
\section{Introduction}
\begin{figure}[t!]
    \includegraphics[width=0.48\textwidth]{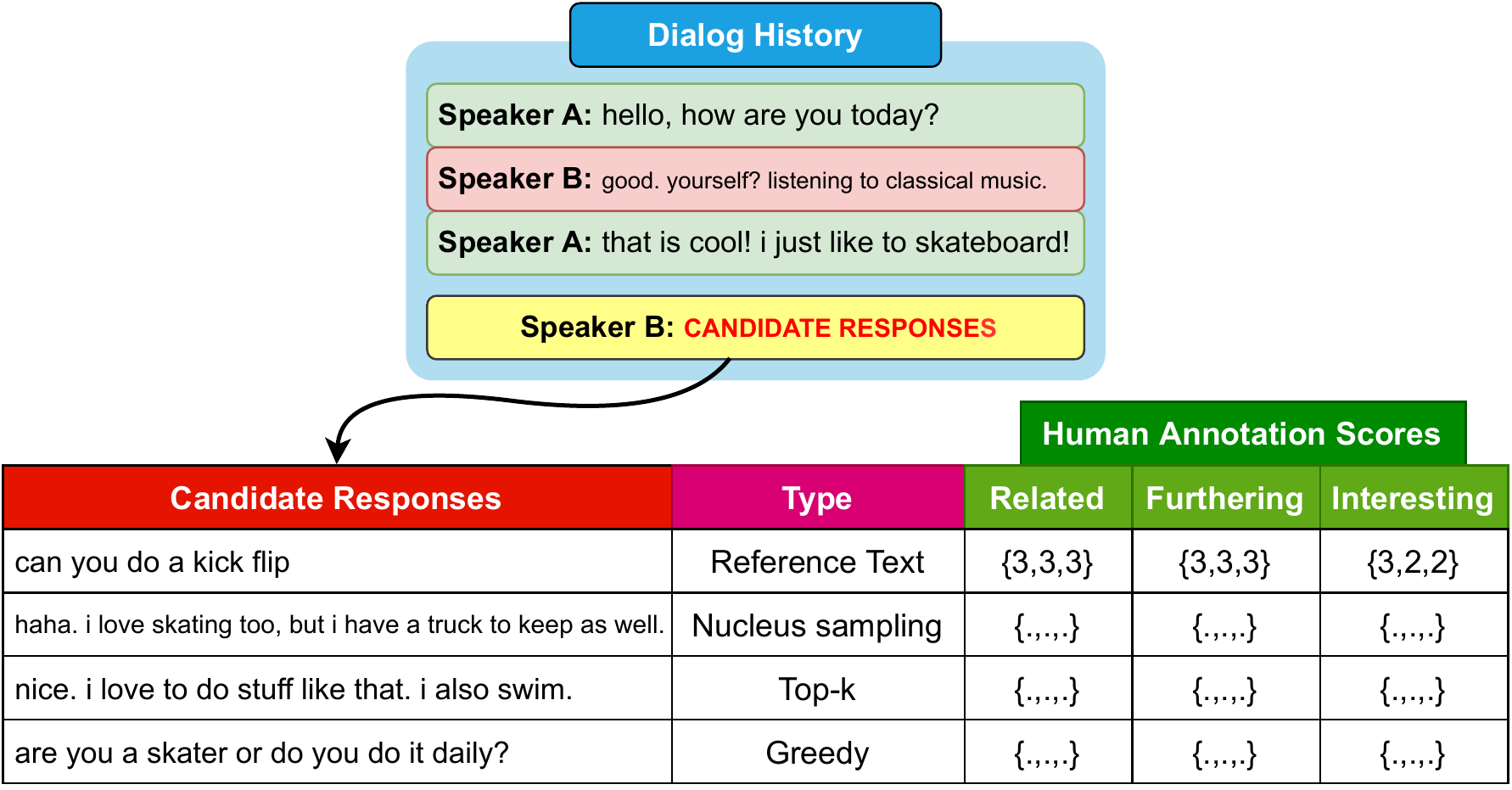}
    \caption{Our dataset contains 4 candidate responses for every dialog history, along with human annotations for 3 qualitative measures.}
    \label{fig:dataset_teaser}
\end{figure}

 The Uniform Information Density (UID) hypothesis states that humans distribute information in their utterances evenly for optimal communication \citep{jaeger2010redundancy, fenk_uid}. Consequently, language generation has benefitted from UID-based objectives and regularization \citep{meister2022typical, regularizer}. Specifically, \citet{meister-etal-2020-beam} argued that UID can be optimized for machine translation using beam search.  Yet, the effect of different decoding algorithms on information density distributions of generated text are unknown, as is UID's broader role in neural response generation in the special case of dialogue models. Here, we investigate (i) if different decoding algorithms follow the UID principle, and (ii) if following the UID principle is beneficial for dialogue response generation, and (iii) collect human annotations of qualitative measures for multiple candidate responses to dialog histories generated using different decoding algorithms (Figure \ref{fig:dataset_teaser}) to study the relationship of dialog response quality and UID. We operationalize UID as the variance of surprisal and measure its correlation with automatic metrics (e.g., BLEU, METEOR, BERTScore) as well as human judgments on qualitative measures of response quality and find that adherence to UID correlates negatively with human judgments when the responses have very low/high surprisal.

 \begin{figure*}[ht]
 \centering
  \includegraphics[width=\textwidth]{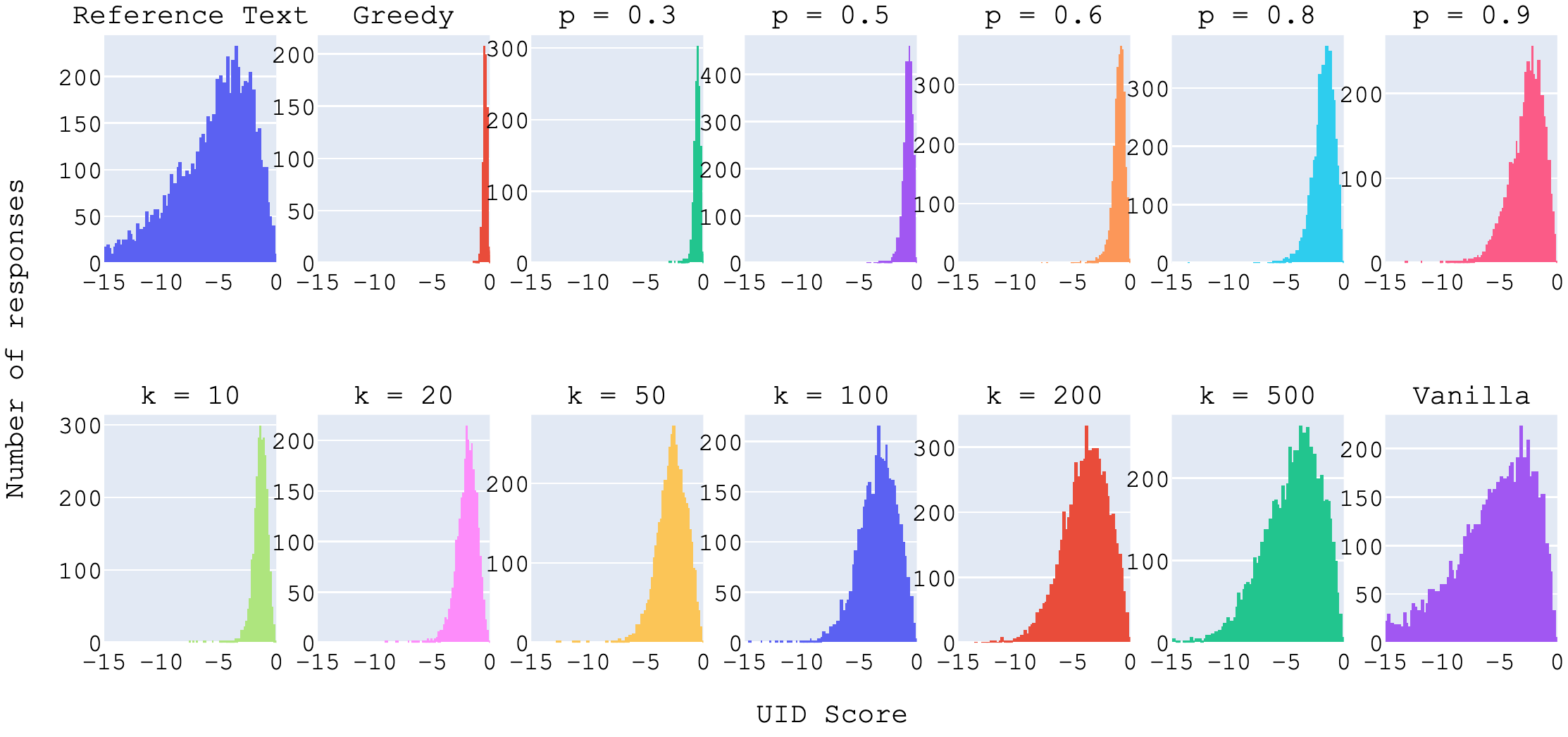}
  \caption{Histogram of \textbf{UID Scores} of responses generated using different decoding algorithms. The farther the UID score from $0$, the less uniform or more non-uniform the response. Human-generated reference text (left-top) has a higher frequency of non-uniform responses as compared to any model setting as can be seen from the wider spread of scores away from $0$. Also, as the values of \textit{p} and \textit{k} increase \textit{(left to right)}, the information density distribution slowly approaches reference text-like non-uniformity.}
  \label{fig:uid}
\end{figure*}

\begin{figure*}[t!]
    \includegraphics[width=\textwidth]{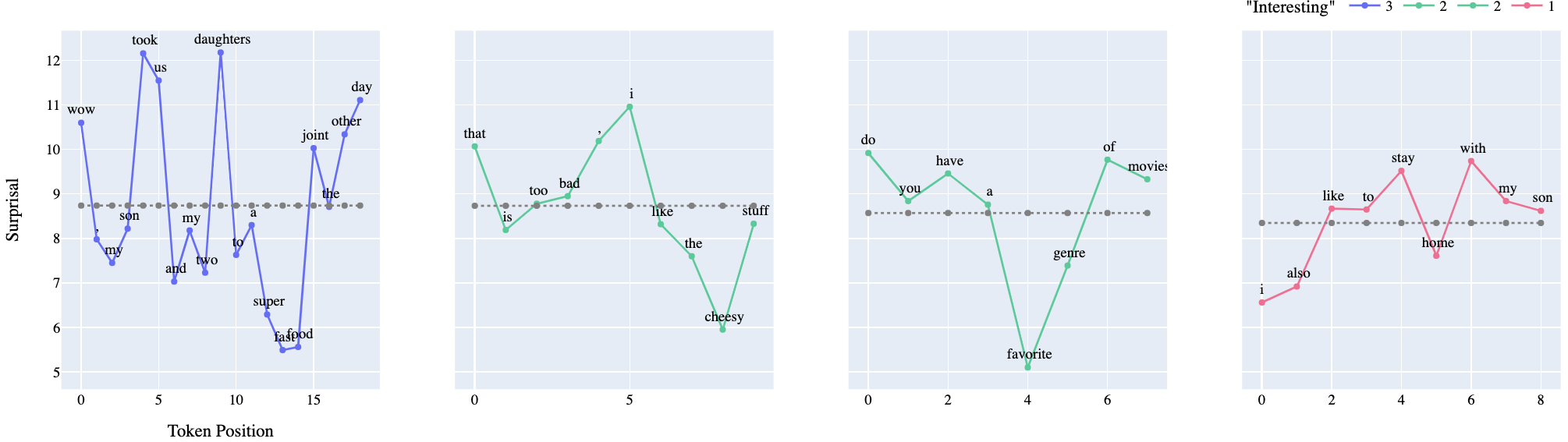}
    \caption{Surprisal at every token in candidate responses to the same dialog history, color-coded with human annotated \textbf{interesting} scores. Plots (\textit{left to right}) are arranged in increasing order of uniformity (i.e. variance along y-axis). Less uniform the surprisal (left-most), better the score.}
    \label{fig:sent_example}
\end{figure*}

\paragraph{Language production in humans.}Spreading information content evenly in utterances is a marker of optimally strategized responses, and humans follow this UID principle as a means to state their thoughts clearly and to make themselves intelligible \citep{frank2008speaking, levy2007speakers}. The probability of a sentence has been associated with the cognitive load it incurs \cite{hale2003information}. As a means to avoid salient variations in the information content (surprisal, i.e., negative log probability) of responses, speakers maintain UID through linguistic choices such as that at the  phonetic \citep{aylett2004smooth}, syntactic \citep{jaeger2010redundancy} and lexical level \citep{mahowald2013info}.

\paragraph{Response generation in machines.} While large-scale pre-trained language models provide a rich prior for dialogue response generation, the choice of decoding algorithm used at the time of generation is crucial for the quality of generated responses \citep{Holtzman2020The, zhang-etal-2021-trading, nadeem-etal-2020-systematic, golovanov-etal-2019-large, oluwatobi-mueller-2020-dlgnet}. While vanilla sampling often tends to produce incoherent text, greedy decoding leads to safe and repetitive responses. More recently, top-\textit{p}/nucleus \citep{Holtzman2020The} and top-\textit{k} sampling \citep{fan-etal-2018-hierarchical} are used to tune values of \textit{p/k} to balance the diversity-quality trade-off \citep{zhang-etal-2021-trading, mutual_info_objective}.

\paragraph{The UID principle and decoding algorithms.} Both the UID principle and decoding algorithms can be seen as guiding mechanisms for dialogue response production in humans and generation in  machines, respectively. UID's role in machine-generated dialogue is not well understood, with previous work mainly focused on machine translation and language modeling \citep{regularizer, meister2021revisiting, meister-etal-2020-beam}. To address this gap, we present a comparative study of decoding methods to develop a deeper understanding of the role of UID in dialogue response generation. 
\section{Experimental Details}
\begin{table*}
\centering
\resizebox{\textwidth}{!}{
\begin{tabular}[width=\textwidth]{c|cccccccc}
& \multicolumn{8}{c}{\textbf{Pearson's \textit{r} between UID score and automatic metrics}} \\
\textbf{Generation Type} & \textbf{Length} & \textbf{BLEU}  & \textbf{chrF} & \textbf{METEOR} & \textbf{BertScore} & \textbf{BLEURT} & \textbf{RoBERTa} & \textbf{SacreBLEU}\\
\hline
\textit{p} = 0.3         & -.10   & .00 & .14          & .12  & .17      & .17   & 0.19                                                            & .13      \\
\textit{p} = 0.5         & -.05 & .03  & .13       & .10 & .18      & .17   & \textbf{.2}                                                             & .15      \\
\textit{p} = 0.6         & -.04 & .06  & .14        & .13  & .01      & .06   & .01                                                            & .00         \\
\textit{p} = 0.8         & -.10 & .03  & .06          & .05  & .18      & .16   & \textbf{.2}                                                             & .15      \\
\textit{p} = 0.9         & -.11 & -.00 & .03        & .04  & .16      & .15   & .19                                                            & .14      \\
Greedy          & -.14 & .01   & .14          & .13  & .06      & .05   & .06                                                            & .06      \\
\textit{k} = 10          & -.04 & .15  & .03        & .05  & .07      & .08   & .07                                                            & .07      \\
\textit{k} = 20          & -.05 & .14  & .05          & .06  & .05      & .04   & .06                                                            & .04      \\
\textit{k} = 50          & -.09 & .01  & .03        & .03  & .06      & .03   & .03                                                            & .05      \\
\textit{k} = 100         & -.07 & .04  & .00       & .02  & .11      & .08   & .08                                                            & .08      \\
\textit{k} = 200         & -.12 & .03  & .02       & .03  & .06      & .06   & .04                                                            & .05      \\
\textit{k} = 500         & -.09 & .02  & .04          & .04  & .10       & .08   & .08                                                            & .08      \\
Vanilla         & -.09 & .01  & -.00 & .00  & .07      & .05   & .05                                                            & .05      \\

\hline
\end{tabular}}
\caption{Pearson's correlation coefficient (\textit{r}) between \textbf{UID score and automatic metrics} of dialog responses generated using different decoding settings. All p-values < 0.05.}
\label{tab:uidcorr}
\end{table*}

\label{sec:experiments}
\subsection{Model \& dataset}  
 We use the fine-tuned GPT-2 \citep{radford2019language} model provided by  HuggingFace and use their data preprocessing and response generation scripts\footnote{\url{https://github.com/huggingface/transfer-learning-conv-ai}}. We used the Persona-Chat \citep{personalizing} data split provided by the ConvAI2 challenge \citep{dinan2020second}\footnote{\url{https://github.com/DeepPavlov/convai/tree/master/2018}}. We then generated responses for 7500 dialogue histories randomly picked from 7801 validation set examples using vanilla, top-\textit{p}, top-\textit{k} sampling and greedy decoding. 
 
\paragraph{Decoding algorithms.} \label{decoding_algos}
\emph{Vanilla sampling} randomly picks the next token from the model's probability distribution, including many long-tail samples. \emph{Top-$k$} samples from the \textit{k} most probable tokens; \emph{Greedy decoding} is Top-$k=1$ decoding, always selecting the most probable next token. \emph{Top-$p$ (Nucleus)} sampling selects the next token from the top $p$ portion of the probability mass.

\subsection{Uniform Information Density score}
We measure UID as the variance of the surprisal (negative log likelihood) of each token in the response \citep{jain-etal-2018-uniform, regularizer, meister-etal-2020-beam}. This measure is able to capture any sudden variations in the surprisal of the tokens in the sentence. UID Score is formulated as follows: the dialogue model learns a conditional probability \textit{p} parameterized by $\theta$ to predict the next token ($y_{t}$) in the sentence. The surprisal ($u$) of the next token $y_{t}$ is, 
\newcommand\eqdef{\ensuremath{\stackrel{\rm def}{=}}} 
\begin{align} 
    u(y_{t}) = - \log  (p_{ \theta } (y | x, y<t)),
\end{align}    
for $t \geq 1$ where $y_{0} = <EOS>$, $t$ = time step, and $x$ = dialogue context. Higher the surprisal, lower its probability and vice-versa. Thus, surprisal indicates how unexpected or surprising a token is in a given context. Average surprisal of a sentence (\textit{y}) is defined as, 

\begin{align}
    \mu(y) = \frac{1}{|y|} \sum_{t}(u(y_{t}))
\end{align} 
Finally, the \textit{UID score} of a sentence (\textit{y}) is defined as the negative normalized variance of the surprisal:
\begin{align}
    \mathrm{UIDscore} (y) =  - \frac{1}{|y|} \sum_{t}(u(y_{t}) -  \mu )^{2}  
\end{align}

From this formulation, a perfectly uniform sentence would have a variance equal to $0$ (i.e. the surprisal of every token in the sentence is equal). Since we take the negative of the variance, the higher the absolute value of UID score, the more non-uniform its information density.


\subsection{Response evaluation}
\paragraph{Automatic metrics.} We measure the quality of responses using length (number of tokens), BLEU\footnote{\label{nltk}\url{https://github.com/nltk/nltk/tree/develop/nltk/translate}} \citep{bleu}, METEOR\textsuperscript{\ref{nltk}} \citep{meteor}, character level F-score (chrF)\textsuperscript{\ref{nltk}} \citep{chrf}, BLEURT\footnote{\label{metrics}\url{https://github.com/huggingface/datasets/tree/master/metrics}} \citep{bleurt}, a RoBERTa \citep{roberta} based text similarity score\footnote{\url{https://github.com/UKPLab/sentence-transformers/blob/master/docs/usage/semantic_textual_similarity.md}} \citep{sts}, BERTscore\textsuperscript{\ref{metrics}} \citep{bertscore} and SacreBLEU\textsuperscript{\ref{metrics}} \citep{sacrebleu}.

\paragraph{Human evaluation.} To study the effect of adherence to UID on the perceived quality of generated responses beyond n-gram, reference-based and learned automatic metrics, we collected human judgments along 3 measures -- \textbf{related} (to the dialogue history), \textbf{furthering} (if a response keeps the conversation going/is encouraging for the dialogue partner) and \textbf{interesting} (if the response provides engaging/new information). We provide screenshots of the task interface (Figure \ref{fig:mturk}), instructions (Figure \ref{fig:mturkex}) and details about the MTurk study design in Appendix \ref{sec:mturk}.
\\

\section{Findings}
 
\subsection{Information density of model responses} We plot the histograms of UID scores computed for all of the generated responses in Figure \ref{fig:uid}. The information densities of human-generated responses have a wider spread than responses produced by the models. Overall, the human-generated reference text has more non-uniform sentences than all model-generated responses. We notice a very high and narrow peak in the case of greedy decoding. This is not surprising as responses sampled using greedy search maximize the probability of the next token (minimize surprisal). Consequently, such responses would have very low surprisal at almost every word, hence lower variance. Vanilla sampling uses the probability distribution learned from the training data, which might be why it is also closer to the validation set (reference text) distribution. With increase in \textit{p} and \textit{k}, we see that the information density distribution spreads across a larger range and includes more non-uniform responses, slowly approaching that of the reference text. \label{sec:hist_results}

\begin{table}[ht!]
\resizebox{0.5\textwidth}{!}{
\begin{tabular}{cc|ccc}
& &   \multicolumn{3}{c}{\textbf{Pearson's \textit{r} between}} \\
& & \multicolumn{3}{c}{\textbf{UID score and qualitative metrics}} \\
\textbf{Surprisal interval} & \textbf{n}   & \textbf{Related}              & \textbf{Furthering}  & \textbf{Interesting}          \\
\hline
& & & & \\
(0.8, 1.2)                                                  & 24  & .17        & -.03       & \textbf{-.30$^{\ast}$}        \\
(1.2, 1.6)                                                  & 64  & .12         & .08       & -.13     \\
(1.6, 2.0)                                                  & 91  & .05         & \textbf{-.23$^{\ast}$}      & -.07        \\
(2.0, 2.4)                                                  & 109 & -.04        & -.13       & -.00       \\
(2.4, 2.8)                                                  & 111 & -.06        & \textbf{-.21$^{\ast}$}       & -.05        \\
(2.8, 3.2)                                                  & 105 & -.02       & .01         & -.10       \\
(3.2, 3.6)                                                  & 99  & \textbf{-.23$^{\ast}$}      & -.10    & .19         \\
(3.6, 4.0)                                                  & 66  & .03         & -.05        & -.09       \\
(4.0, 4.4)                                                  & 42  & -.33       & -.22       & -.09     \\
(4.4, 4.8)                                                  & 24  & -.14        & \textbf{-.61$^{\ast}$}      & .04        \\
(4.8, 5.2)                                                  & 12  & -.33       & -.14     & \textbf{-.54$^{\ast}$}        \\
(5.2, 5.6)                                                  & 13  & \textbf{-.98$^{\ast}$}       & -.64      & -.38      \\
\\
\hline
\end{tabular}}
\caption{Pearson's \textit{r} between \textbf{UID score and and human judgments} of qualitative measures for dialog responses bucketed by surprisal [Surprisal interval = the ranges of surprisal values used for bucketing responses, n = number of responses in each surprisal interval, $^{\ast}$p-value < .05]}
\label{tab:humaneval}
\end{table}

\subsection{UID score \& automatic metrics}
We present the correlation between UID scores and automatic metrics calculated for the generated dialogue responses in Table \ref{tab:uidcorr}. UID scores have a weak correlation with RoBERTa-based similary scores for two settings of nucleus sampling. Other than that, UID scores are not correlated with automatic metrics of response generation. We take this to be an indication that if UID scores do capture any aspect of response quality, it goes beyond what is measured by such metrics and might provide for a better evaluation criteria.


\subsection{UID score \& human Judgments} 
Motivated by the fact that UID score is derived from surprisal, we test if surprisal is a confounding factor and find that, indeed, UID scores were highly correlated with average surprisal (Table \ref{tab:suprisal_uid}). To tease apart the effect of UID scores on response quality, we controlled for surprisal by grouping or bucketing responses into 12 intervals of surprisals (within a range of 0.4 units as shown in the first column on Table \ref{tab:humaneval}). Within these intervals, surprisal had no correlation with generation quality (Table \ref{tab:corr_human_surp}). Once we control for surprisal i.e. analyse dialog responses with similar surprisals but varying UID scores, we observe that UID scores negatively correlate with human judgments, to varying degrees of strength, for responses in very low or high surprisal intervals (see Table \ref{tab:humaneval}). Thus, for the extremities of the surprisal range, UID scores indicate that better rated responses are non-uniform.

\section{Discussion}
Contrary to our expectations, we find non-uniformity to be a more desirable property in machine-generated responses. Overall, UID scores and surprisal do not correlate with human judgments (Table \ref{tab:judgments}). But when controlled for surprisal, we observe that UID score is correlated with human judgments for certain intervals (examples in Figure \ref{fig:sent_example} and Table \ref{tab:mturk_examples}). Our results suggest that optimizing UID to generate uniform text might not be the right objective for regularizing decoding algorithms. Instead we find that non-uniform information density could be a potential solution to the ``likelihood trap" problem according to which models generate lower quality text (as per human judgments) when sampling from the extremities of their likelihood space \cite{trading}. Consequently, we suggest that decoding algorithms be tuned to follow the information density patterns of human-generated non-uniform data when generating responses outside of the ``safe'' likelihood range as a means to generate higher quality responses across the entire likelihood space.

\section{Limitations}
While we present a study of multiple decoding settings, we generate all machine responses using the same transformers based model architecture. Thus, the presented work does not yet explore individual differences between different model architectures. Additionally, due to limited resources we were not able to collect large-scale human annotations across multiple corpora and acknowledge the same as part of future efforts.  

\section{Ethical considerations}
In this work, we collected human annotations on dialogue response quality using MTurk. Each HIT in our MTurk study contained one dialogue history and four candidate responses. The annotators could read the history and rate the responses that followed using mouse clicks on their response choices. We provided an additional feedback field for annotators to write comments in. We received very positive feedback on the task from all the annotators who used this feature. There were no restrictions on the minimum or maximum number of examples the annotators had to rate. From a pilot study on MTurk, we found the average time to complete one HIT to be slightly under 2.5 minutes. After considering the average time required and the task difficulty (expressed to be clearly and easily understood by annotators in their comments) we set the payment amount to \$0.5 per HIT for an hourly rate of about \$12 per hour.
\bibliography{anthology,custom}
\bibliographystyle{acl_natbib}

\clearpage
\appendix
\begin{table}[h]
\begin{tabular}[0.5\textwidth]{cc}
\hline
\textbf{Generation Type} &\textbf{Pearson's \textit{r}} \\
\hline
Reference Text  & -.69      \\
Greedy          & -.23      \\
\textit{p} = 0.3         & -.43   \\
\textit{p} = 0.5         & -.50      \\
\textit{p} = 0.6         & -.56  \\
\textit{p} = 0.8         & -.65     \\
\textit{p} = 0.9         & -.68      \\
\textit{k} = 10          & -.40   \\
\textit{k} = 20          & -.45      \\
\textit{k} = 50          & -.56     \\
\textit{k} = 100         & -.63     \\
\textit{k} = 200         & -.65     \\
\textit{k} = 500         & -.69     \\
Vanilla         & -.74     \\
\hline
\end{tabular}
\caption{Pearson's correlation coefficient (\textit{r}) \textbf{between UID score and average sentence surprisal} (all \textit{p} < 0.01)}
\label{tab:suprisal_uid}
\end{table}

\begin{figure}[h]
\includegraphics[width=0.5\textwidth]{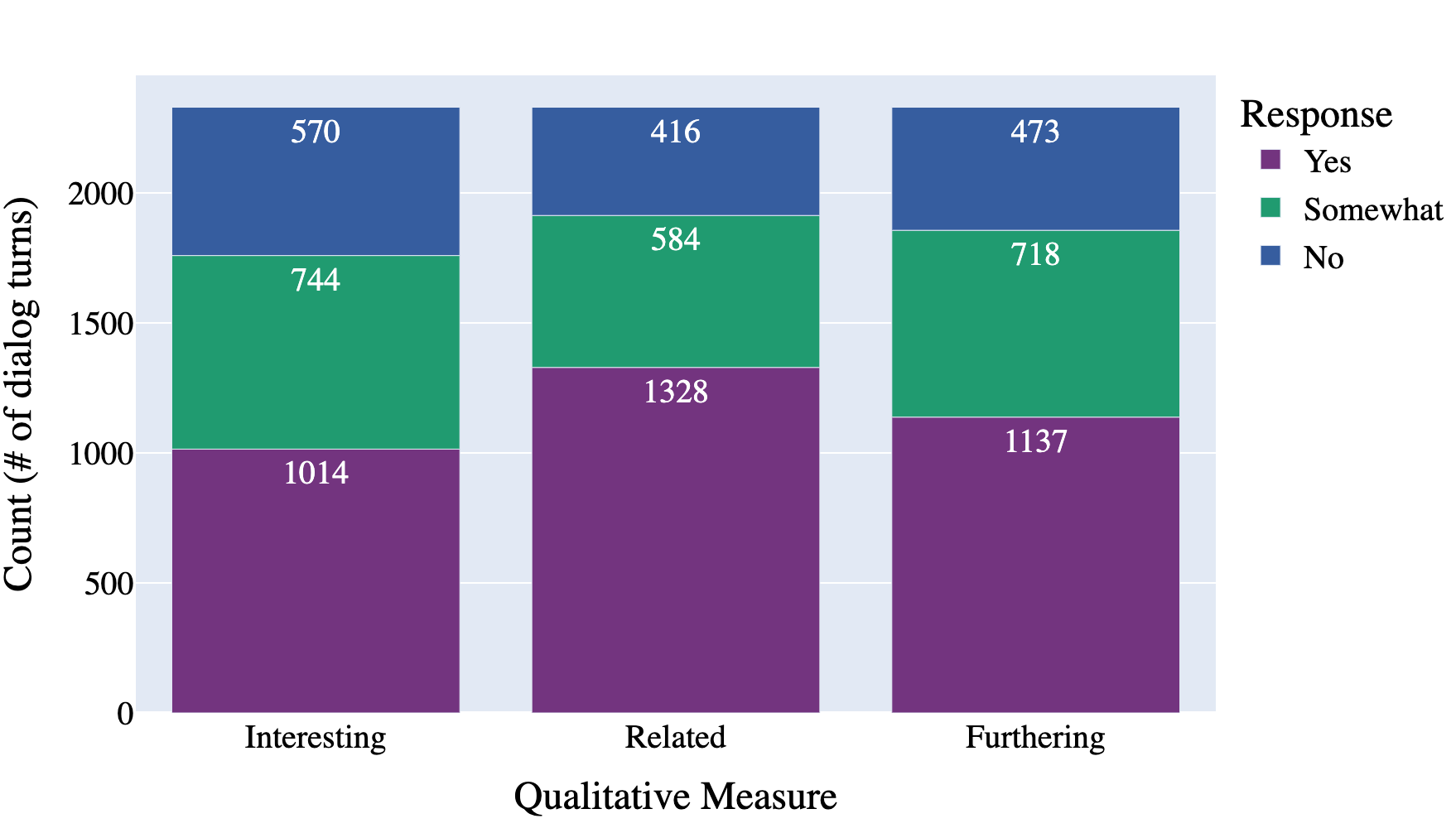}
\caption{Frequency of responses (Yes/Somewhat/No) for each qualitative measure in our human annotated dataset.}
\label{fig:human_data_dist}
\end{figure}

\begin{table}[h]
\centering
\begin{tabular}{c|cc}
& \multicolumn{2}{c}{\textbf{Pearson's \textit{r}}} \\
\textbf{Quality} & \textbf{UID Score} & \textbf{Surprisal}   \\
\hline
Related     & .01  & \textbf{-.13$^{\ast}$}    \\
Furthering      & .03  & \textbf{-.10$^{\ast}$} \\
Interesting & -.04 & -.01 \\
\hline
\end{tabular}
\caption{Pearson's correlation coefficient (\textit{r}) of \textbf{UID score and surprisal with human judgments of qualitative metrics} ($^{\ast}$\textit{p}<0.01)}
\label{tab:judgments}
\end{table}
\begin{figure*}[h!]
 \centering
  \includegraphics[width=\textwidth]{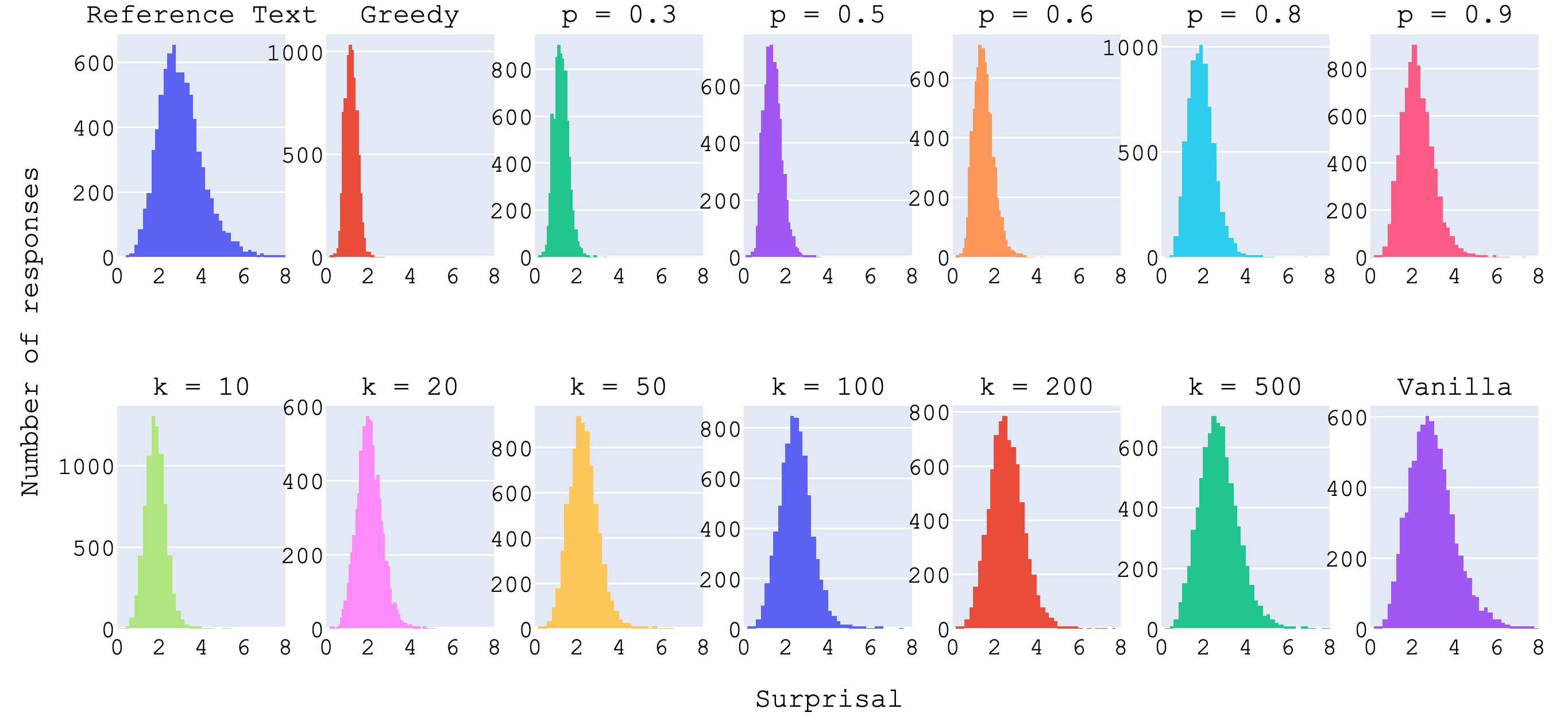}
  \caption{Histograms of \textbf{average sentence surprisal} for responses generated using different decoding settings and human-generated reference text (left-top).}
  \label{fig:surp}
\end{figure*}

\begin{table}[h]
\resizebox{0.5\textwidth}{!}{
\begin{tabular}[0.5\textwidth]{cc|ccc}
& &   \multicolumn{3}{c}{\textbf{Pearson's \textit{r}}} \\
\textbf{Surprisal  interval} & \textbf{n}   & \textbf{Related}  & \textbf{Furthering}  & \textbf{Interesting}   \\
\hline
\\
(0.8,1.2)       & 24  & -.03     & -.04      & -.00    \\
(1.2,1.6)       & 64  & -.10     & -.16     & .08    \\
(1.6,2.0)       & 91  & .05     & .14    & .10   \\
(2.0,2.4)       & 109 & -.14    & -.08     & \textbf{-.27$^{\ast}$}\\
(2.4,2.8)       & 111 & -.12    & .05      & .09     \\
(2.8,3.2)       & 105 & -.02   & .06      & -.00    \\
(3.2,3.6)       & 99  & -.13   & .12       & .01  \\
(3.6,4.0)       & 66  & .02     & -.06    & .06     \\
(4.0,4.4)       & 42  & -.01    & -.00     & .06     \\
(4.4,4.8)       & 24  & .20    & .34     & .23    \\
(4.8,5.2)       & 12  & -.13      & -.37     & -.12     \\
(5.2,5.6)       & 13  & .60         & .83      & .76 \\
\hline
\end{tabular}}
\caption{Pearson's \textit{r} between \textbf{surprisal and human judgments} of qualitative measures for dialog responses bucketed by surprisal [Surprisal interval = the ranges of surprisal values used for bucketing responses, n = number of responses in each surprisal interval, $^{\ast}$p-value < .05]}
\label{tab:corr_human_surp}
\end{table}

\section{Human evaluation study details} \label{sec:mturk}
Raters were selected based on the criteria that they be located in the US, and had attempted a minimum of 500 HITS at an accepted work rate greater than 97\% on MTurk.
 We asked raters on MTurk to answer if a candidate response satisfied each of the qualitative measures (interesting, furthering and related) and gave them three response options: "Yes", "Somewhat" and "No". In a pilot study of $360$ responses, we also included a measure for fluency. All of the responses were rated ``Yes" by majority vote and we removed this measure from further analysis as all the generations in this study were fluent as indicated by the pilot study and from our observation. For correlation calculations, we assign integer score values to each of the three response options as $3$ for "Yes", $2$ for "Somewhat" and $1$ for "No". Thus, the higher the score, the better the response is rated. Following the pilot study, for 194 dialogue histories, we showed the raters 4 candidate dialogue responses (total of 776 dialogue responses) and collected ratings on all *3* measures from *3* raters per dialogue history. In all, we obtained a total of 776*3, i.e., 2328 total response-rating pairs. To calculate the score for each response along every measure, we take the mean of all ratings as the score. For cases where at least 2 out of 3 raters agree, we take majority vote  as the final score. This constituted (2018 out of 2328) 86.68\% of all the ratings collected. We show the overall distribution of qualitative scores for all the response-rating pairs in Figure \ref{fig:human_data_dist}. We verified the rater responses by checking if they were rating human-generated responses highly as those came from a trusted source (Persona-Chat). We also manually inspected a random subset of dialog history-candidate response sets and found the results to be in accordance with our intuitions.

\begin{figure*}
\centering
    \begin{subfigure}{\textwidth}
        \includegraphics[width=\textwidth]{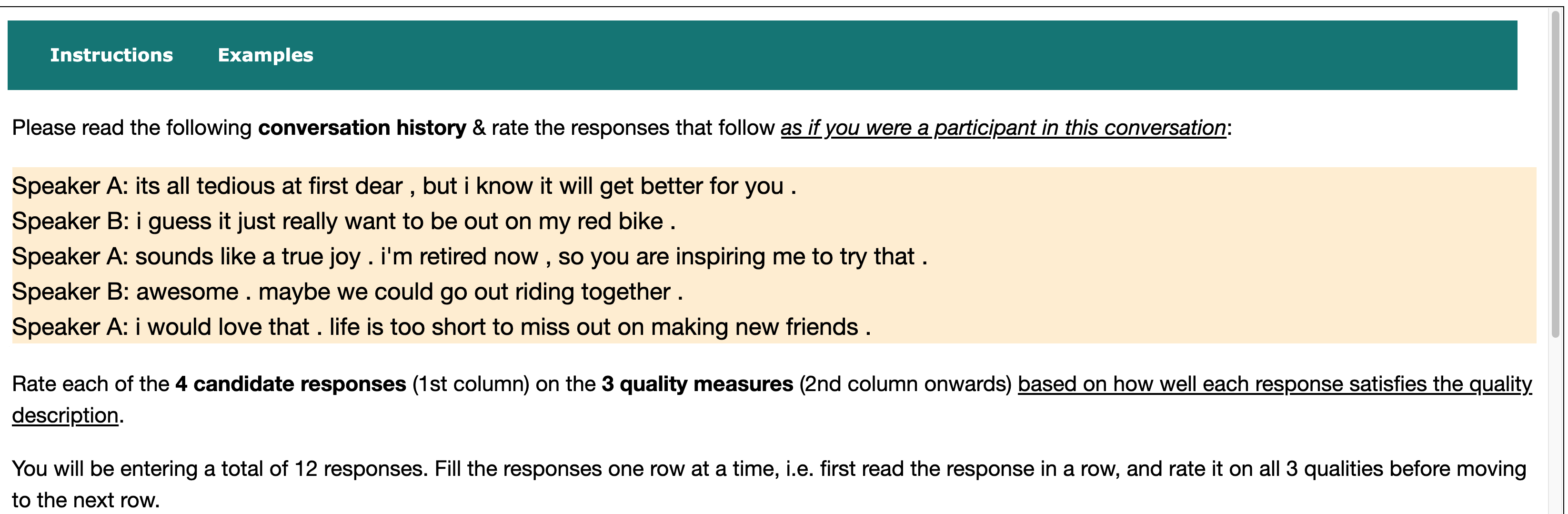}
    \end{subfigure}
    \\
    \begin{subfigure}{\textwidth}
        \includegraphics[width=\textwidth]{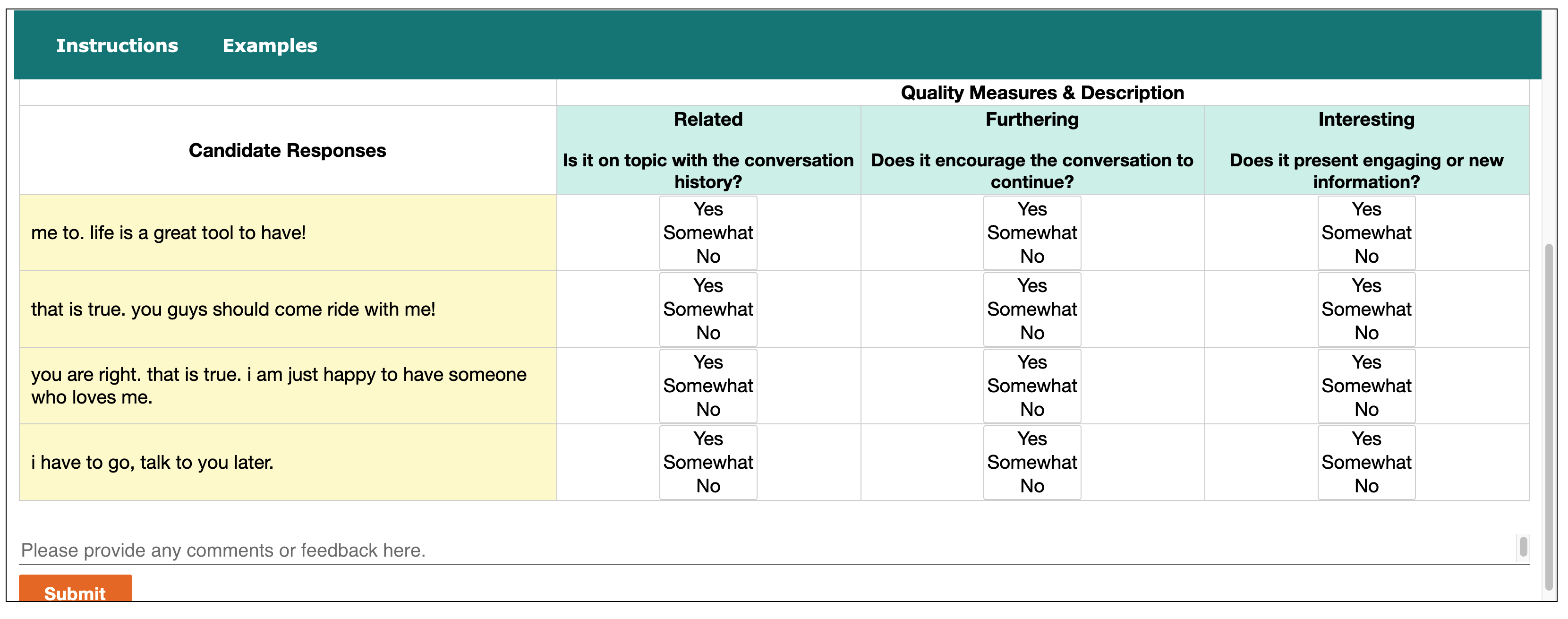}
    \end{subfigure}
\caption{Screenshots of our MTurk study interface for collecting human judgments on 4 candidate responses per dialogue history, along 3 quality measures.}
\label{fig:mturk}
\end{figure*}

\begin{figure*}
 \begin{subfigure}{\textwidth}
        \includegraphics[width=\textwidth]{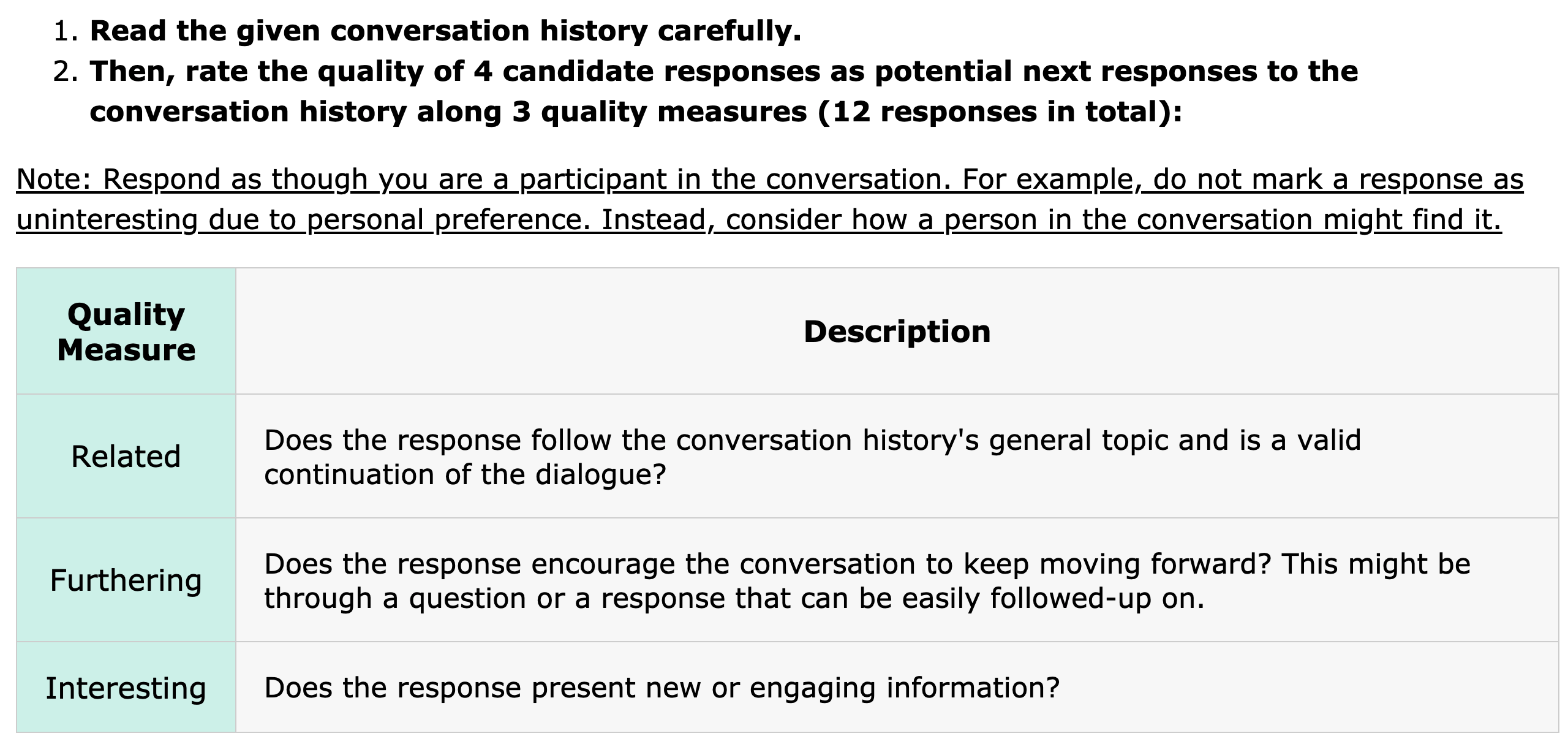}
        \caption{Detailed instructions that MTurk raters could expand at any time.}
    \end{subfigure}
    \\
    \\
    \begin{subfigure}{\textwidth}
        \includegraphics[width=\textwidth]{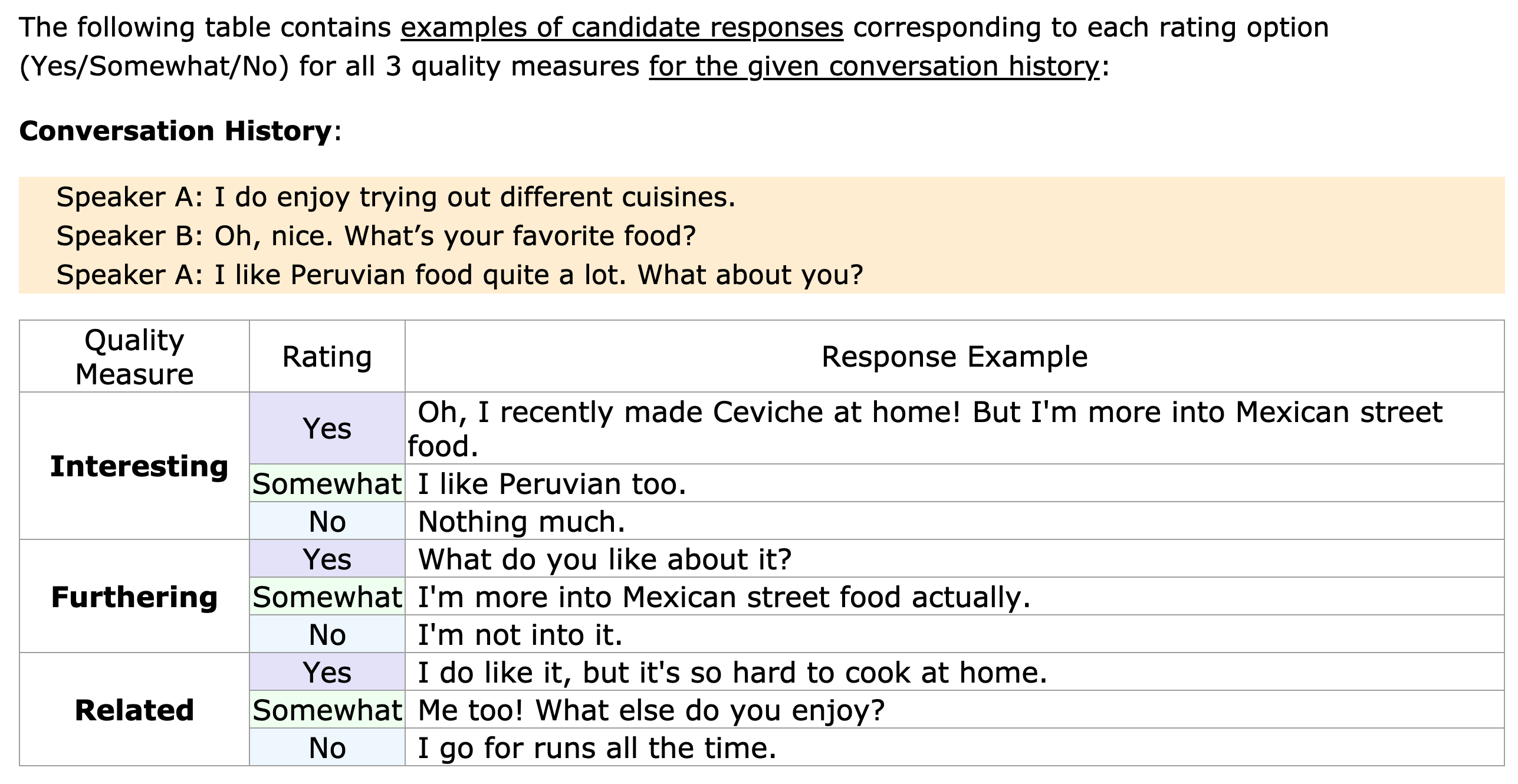}
        \caption{Examples responses for each measure and rating category shown to MTurk raters.}
    \end{subfigure}
\caption{Instructions and examples from MTurk study.}
\label{fig:mturkex}
\end{figure*}

\begin{table*}[t!]
\small
\centering
\resizebox{\textwidth}{!}{ 
\begin{tabular}{llcc}
\multicolumn{4}{c}{\underline{\textbf{Dialogue history}}} \\
 & & & \\

\multicolumn{4}{c}{\textbf{Speaker A: hi, i m susie. how are you?}}\\
 & & & \\
\textbf{Response type} &  \textbf{Response text} &  \textbf{UID Score}  &  \textbf{Interesting}  \\
 & & & \\
Reference Text  & i'm good. just got back from some volunteering. how are you doing? & -1.99  & 3  \\
Candidate 1 & hi. i am very good. just enjoying my favorite pastime. how are you?   & -1.43  & 2   \\
Candidate 2 & i am great! i volunteer at a soup kitchen and watch movies with my kids. & -1.19  & 2  \\
Candidate 3  & i'm doing well. how are you? & -0.18 & 1 \\
 & & & \\
 \hline
 & & & \\
 \multicolumn{4}{c}{\begin{tabular}[c]{@{}l@{}}
\textbf{Speaker A: that's cool. my dad made us italian food tonight.} \\
\textbf{Speaker B: oh nice, i love italian food. my favorite is the garlic bread. what is yours?}\\
\textbf{Speaker A: pasta, my son loves pizza though!}\\
\textbf{Speaker B: i like that too, have to eat lots of carbs for my training.}
\end{tabular}}\\
 & & & \\
\textbf{Response type} & \textbf{Response text} &  \textbf{UID Score}  &  \textbf{Interesting}\\
& & & \\
Candidate 1    & wow, my son took us and my two daughters to a super fast food joint the other day & -4.83 & 3 \\
Candidate 2    & that is too bad, i like the cheesy stuff.                                         & -2.08 & 2 \\
Reference Text & do you have a favorite genre of movies?                                           & -1.73 & 2 \\
Candidate 3    & i also like to stay home with my son.                                             & -0.88 & 1\\
 & & & \\
 \hline
 & & & \\ 
\multicolumn{4}{c}{\begin{tabular}[c]{@{}l@{}}
\textbf{Speaker A: hello i hope your sunday is great , what is your favorite kind of music?}\\
\textbf{Speaker B: hey there . been a relaxed sunday . yours ? music eclectic.}\\
\textbf{Speaker A: my sunday has been exciting ! i enjoy death metal.}\\
\textbf{Speaker B: death metal . cool . i spent the morning volunteering.}\\
\textbf{Speaker A: i volunteer too , at the local pool to be a swim coach.} \end{tabular}}\\
 & & & \\
\textbf{Response type} &  \textbf{Response text} &  \textbf{UID Score}  &  \textbf{Furthering}  \\
& & & \\
Candidate 1    & nice. such a nice day. how long have you been coaching?          & -9.99 & 3 \\
Reference Text & that is great! we both volunteer! mine is rescuing bunnies.      & -7.59 & 2 \\
Candidate 2    & that's cool. i donate my pay to the local zoo. humane societies. & -3.89 & 2 \\
Candidate 3    & sick sick. beautiful color, navy blue is my favorite.            & -2.79 & 1\\
 & & & \\
 \hline
 
 & & & \\
\multicolumn{4}{c}{\begin{tabular}[c]{@{}l@{}}
\textbf{Speaker A: have you heard about the juggalos? weird.}\end{tabular}} \\
 & & & \\
\textbf{Response type} &  \textbf{Response text} &  \textbf{UID Score}  &  \textbf{Furthering}  \\
& & & \\
Reference Text & what are those? do they juggle balls?     & -6.88 & 3 \\
Candidate 1    & no i have not. i like that band myself    & -5.19 & 2 \\
Candidate 2    & we will not tell. do you like comics?     & -4.27 & 2 \\
Candidate 3    & i have not heard of them. i've dolls too. & -2.8  & 1 \\
 & & & \\
\hline
 & & & \\
\multicolumn{4}{c}{\begin{tabular}[c]{@{}l@{}}
\textbf{Speaker A: hello, how are you today?}\\
\textbf{Speaker B: good. yourself? listening to classical music.}\\
\textbf{Speaker A: that is cool! i just like to skateboard!}\\
\end{tabular}}\\
  & & & \\
\textbf{Response type} &  \textbf{Response text} &  \textbf{UID Score}  &  \textbf{Related}\\
& & & \\
Reference Text & can you do a kick flip? & -14.11 & 3 \\
Candidate 1    & haha. i love skating too, but i have a truck to keep as well. & -11.28 & 3 \\
Candidate 2    & nice. i love to do stuff like that. i also swim.  & -10.25 & 3 \\
Candidate 3  & are you a skater or do you do it daily?  & -9.05  & 2\\
& & & \\

\hline
 & & & \\
\multicolumn{4}{c}{\begin{tabular}[c]{@{}l@{}}
\textbf{Speaker A: i hope it is a healthy snack. i am doing well, just moved here.}\\
\textbf{Speaker B: it is. i am vegan. just moved here too! where from?}\\
\textbf{Speaker A: i too am vegan from germany. have you seen lafer! lichter! lecker! on tv?}\\
\textbf{Speaker B: oh yes! i love shows like that and watched it back home a lot.}\\
\end{tabular}}\\
 & & & \\
\textbf{Response type} &  \textbf{Response text} &  \textbf{UID Score}  &  \textbf{Related}\\
& & & \\
Reference Text & they do amazing things with the hummus. where is back home for you? & -6.86 & 3 \\
Candidate 1    & you must have a lot of fun watching them.                           & -4.53 & 2 \\
Candidate 2    & they have the best new vegan cookbooks, but i am more adventurous.  & -2.93 & 3 \\
Candidate 3    & i do love the sky diving, too! i have seen the first few seasons.   & -2.47 & 1 \\

\end{tabular}}
\caption{Examples of dialogue histories followed by 4 response candidates arranged by increasing UID score i.e. from more non-uniform to uniform responses and their corresponding human judgment scores.}
\label{tab:mturk_examples}
\end{table*}

\end{document}